\documentclass[conference]{IEEEtran}
\IEEEoverridecommandlockouts
\IEEEpubid{\makebox[\columnwidth]{%
\parbox{\columnwidth}{\footnotesize
\copyright~2026 IEEE. Personal use of this material is permitted.
Permission from IEEE must be obtained for all other uses, in any current
or future media, including reprinting/republishing this material for
advertising or promotional purposes, creating new collective works, for
resale or redistribution to servers or lists, or reuse of any copyrighted
component of this work in other works.}%
\hfill} \hspace{\columnsep}\makebox[\columnwidth]{}}

\usepackage{cite}
\usepackage{amsmath,amssymb,amsfonts}
\usepackage{algorithmic}
\usepackage{graphicx}
\usepackage{microtype}
\usepackage{textcomp}
\usepackage{xcolor}
\def\BibTeX{{\rm B\kern-.05em{\sc i\kern-.025em b}\kern-.08em
    T\kern-.1667em\lower.7ex\hbox{E}\kern-.125emX}}
\begin{document}

\title{Dynamic Evaluation of Classical and Control-Aware Optimal Trajectory Planning in Robot Manipulators\\

}

\author{\IEEEauthorblockN{Sachintha Bhanuka Dayawansa}
\IEEEauthorblockA{\textit{Department of Electronic and Telecomm. Engineering} \\
\textit{University of Moratuwa}\\
Katubedda, Sri Lanka \\
dayawansasb.22@uom.lk}
\and
\IEEEauthorblockN{Sudath Rohan Munasinghe}
\IEEEauthorblockA{\textit{Department of Electronic and Telecomm. Engineering} \\
\textit{University of Moratuwa}\\
Katubedda, Sri Lanka \\
University of Tartu,  Estonia\\
rohan@uom.lk}

}

\maketitle
\IEEEpubidadjcol

\begin{abstract}
Trajectory planning strongly influences tracking accuracy, actuator demand, and overall execution behavior in robotic manipulators. Classical planners such as cubic, quintic, and trapezoidal profiles are widely used for their simplicity and smoothness, yet they remain purely kinematic and ignore system dynamics and control effort during trajectory generation. As a result, nominally smooth trajectories can lead to inefficient nonlinear execution and increased corrective control action.
This paper presents a control-aware optimal trajectory planning framework that explicitly incorporates manipulator dynamics and actuator effort within a finite-horizon formulation. A midpoint linearization strategy is introduced to improve approximation accuracy for large point-to-point motions. In contrast to prior comparisons, the proposed approach enables fair, isolated evaluation of trajectory generation effects under identical closed-loop nonlinear execution conditions. To this end, a unified evaluation framework is developed in which all planners are executed under identical nonlinear dynamics, controller structure, and actuator constraints.
Simulations on a nonlinear simplified UR5 manipulator show that the proposed approach consistently reduces tracking error, corrective torque, and closed-loop execution cost compared to classical methods, achieving substantial reductions in actuator effort and execution cost across all evaluated scenarios, demonstrating that kinematic smoothness alone does not ensure dynamically efficient execution.

\end{abstract}

\begin{IEEEkeywords}
trajectory planning, optimal control, robotic manipulators, nonlinear execution
\end{IEEEkeywords}

\section{Introduction}

Trajectory planning significantly influences tracking accuracy, actuator demand, and overall execution behavior in robotic manipulators. Classical trajectory generation methods such as cubic polynomials, quintic polynomials, and trapezoidal velocity profiles are widely used due to their simplicity, computational efficiency, and smoothness properties \cite{b1}. However, these approaches are fundamentally kinematic and do not explicitly account for manipulator dynamics or actuator control effort during trajectory generation \cite{b1,b2}. Consequently, trajectories that appear smooth in joint space may still produce unfavorable acceleration distributions, elevated corrective actuator demand, and dynamic inefficiencies when executed under nonlinear manipulator dynamics \cite{b3}.

In practical robotic systems, trajectories are executed through feedback controllers operating on nonlinear manipulator dynamics; therefore, trajectory quality cannot be evaluated solely from kinematic smoothness. Even geometrically smooth trajectories may impose substantial corrective burden on the feedback controller, leading to aggressive actuator behavior and increased execution cost. This raises an important question regarding whether widely used classical trajectory planners remain dynamically efficient under realistic nonlinear execution conditions \cite{b4,b5}.

Optimal control methods provide a framework for generating dynamically feasible trajectories by explicitly incorporating system dynamics and actuator effort into the optimization process \cite{b6}. Although advanced nonlinear and receding-horizon optimization approaches have demonstrated strong performance, many introduce significant computational complexity that may be unnecessary for structured industrial manipulation tasks where offline trajectory generation combined with feedback tracking control is sufficient \cite{b7}. In addition, recent actuator-aware trajectory optimization studies suggest that higher-order kinematic smoothness alone does not necessarily correspond to dynamically efficient manipulator execution \cite{b8}.

Based on these observations, this work investigates the nonlinear execution behavior of classical trajectory planners through comparison with a control-aware optimal trajectory planning framework. The proposed approach employs a finite-horizon optimal control formulation that penalizes both state deviation and actuator effort, while a midpoint linearization strategy is used to better approximate nonlinear manipulator dynamics during large point-to-point motions.Furthermore, a unified nonlinear evaluation framework is developed in which all trajectory planners, including cubic, quintic, trapezoidal, and the proposed optimal planner, are executed under identical nonlinear manipulator dynamics, feedforward-assisted PID tracking control, controller gains, and actuator constraints. This enables execution differences to be directly attributed to the trajectory generation method itself. Unlike prior studies, this work provides a unified nonlinear evaluation framework that isolates the effect of trajectory generation on closed-loop execution by enforcing identical dynamics, controller structure, and actuator constraints across all methods.

Simulation studies using a nonlinear 3-DoF UR5 manipulator demonstrate that the proposed control-aware planner achieves lower tracking error, reduced corrective actuator demand, and lower closed-loop execution cost compared to classical trajectory planners. The results further demonstrate that higher-order kinematic smoothness alone does not necessarily correspond to dynamically efficient nonlinear manipulator execution. The proposed planner is not intended as a new LQR/MPC/iLQR solver; rather, it uses a finite-horizon control-effort-aware formulation as an offline trajectory generator and evaluates all planners under the same nonlinear PID-based execution framework.

\section{Related Work}

\subsection{Classical Trajectory Planning}

Classical robotic manipulator trajectory generation commonly relies on kinematic methods such as cubic and quintic polynomial interpolation, trapezoidal velocity profiles, and spline-based formulations due to their simplicity, computational efficiency, and smoothness characteristics \cite{b1}. However, these methods remain fundamentally kinematic and do not explicitly incorporate manipulator dynamics or actuator control effort during trajectory generation \cite{b9}.

\subsection{Optimal Control Methods}

To address the limitations of purely kinematic planners, optimal control-based trajectory generation methods incorporate system dynamics and control-related objectives directly into the optimization process. Approaches such as direct transcription, differential dynamic programming, sequential quadratic optimization, and Model Predictive Control (MPC) have been widely applied for dynamically feasible trajectory generation under system constraints \cite{b6,b10}. By penalizing quantities such as state deviation and actuator effort, these methods can improve execution behavior and reduce excessive control demand \cite{b11}. 
\subsection{Control-Effort-Aware and Dynamically-Aware Trajectory Planning}

Recent research has explored trajectory optimization methods that explicitly consider actuator demand, torque minimization, and dynamic execution characteristics during trajectory generation. Multi-objective optimization approaches incorporating smoothness, execution time, and control effort have shown that higher-order kinematic smoothness alone does not necessarily correspond to dynamically efficient manipulator execution \cite{b9,b13}.

\subsection{Gap Statement}

Although substantial work exists in both classical and optimal trajectory planning, the influence of trajectory generation methods on corrective control effort and nonlinear execution behavior remains insufficiently investigated. Furthermore, many comparative studies employ different controllers, simulation environments, or dynamic models, making it difficult to isolate the effect of the trajectory planner itself. To address these limitations, this work develops a unified nonlinear evaluation framework in which all trajectory planners are executed under identical nonlinear dynamics, controller configurations, and actuator constraints, enabling fair comparison of corrective actuator demand and closed-loop nonlinear execution behavior \cite{b1,b12,b13}.

\section{Problem Formulation}

This work considers control-aware trajectory generation for robotic manipulators under nonlinear dynamics. The objective is to generate dynamically feasible joint-space trajectories that minimize state deviation and actuator control effort while driving the manipulator to a desired terminal configuration.

\subsection{Nonlinear System Dynamics}

The dynamics of an $n$-DoF robotic manipulator are

\begin{equation}
\tau = D(q)\ddot{q} + C(q,\dot{q})\dot{q} + G(q)
\end{equation}

where $q \in \mathbb{R}^{n}$ denotes joint positions and $\tau \in \mathbb{R}^{n}$ denotes actuator torques.

Using gravity compensation,

\begin{equation}
\tau = G(q) + u
\end{equation}

gives

\begin{equation}
\ddot{q} = D(q)^{-1}\left(u - C(q,\dot{q})\dot{q}\right)
\end{equation}

Defining

\begin{equation}
x =
\begin{bmatrix}
q & \dot{q}
\end{bmatrix}^{T}
\end{equation}

the nonlinear dynamics become

\begin{equation}
\dot{x} = f(x,u) =
\begin{bmatrix}
\dot{q}\\
D(q)^{-1}\left(\tau - C(q,\dot{q})\dot{q} - G(q)\right)
\end{bmatrix}
\end{equation}

\subsection{Midpoint Linearization of the Manipulator Dynamics}

The nonlinear dynamics are linearized around the midpoint operating pair $(\bar{x},\bar{u})$ using deviation variables

\begin{equation}
\delta x = x - \bar{x}, \quad \delta u = u - \bar{u}
\end{equation}

yielding the first-order approximation

\begin{equation}
\delta \dot{x}(t) = A_c \delta x(t) + B_c \delta u(t)
\end{equation}

where

\begin{equation}
A_c =
\frac{\partial f}{\partial x}
\bigg|_{(\bar{x},\bar{u})},
\quad
B_c =
\frac{\partial f}{\partial u}
\bigg|_{(\bar{x},\bar{u})}
\end{equation}

The resulting linearized model is

\begin{equation}
A_c(t) =
\begin{bmatrix}
0 & I\\
A_{21}(t) & A_{22}(t)
\end{bmatrix},
\quad
B_c(t) =
\begin{bmatrix}
0\\
D^{-1}(\bar{q}(t))
\end{bmatrix}
\end{equation}

Using Zero-Order Hold discretization with sampling time $T_s$,

\begin{equation}
A_d = e^{A_c(t_k)T_s}
\end{equation}

\begin{equation}
B_d = \int_{0}^{T_s} e^{A_c(t_k)\lambda} d\lambda \, B_c(t_k)
\end{equation}

The trajectory planning problem is then formulated over the control sequence

\begin{equation}
U = \{u_0,u_1,\ldots,u_{N-1}\}
\end{equation}

\section{Control-Aware Optimal Control Framework}

\subsection{Regulation-State Representation}

Define the desired goal state as

\begin{equation}
x_g =
\begin{bmatrix}
q_g & \dot{q}_g
\end{bmatrix}^{T}
\end{equation}

and the regulation-state vector as

\begin{equation}
x_k =
\begin{bmatrix}
q_k - q_g\\
\dot{q}_k - \dot{q}_g
\end{bmatrix}
\end{equation}

Using the discrete-time dynamics,

\begin{equation}
x_{k+1} = A_d x_k + B_d u_k
\end{equation}

the trajectory planning problem is formulated as a finite-horizon regulation problem.

\subsection{Finite-Horizon Prediction Model}

Define the stacked state and control vectors

\begin{equation}
X =
\begin{bmatrix}
x(1)\\
x(2)\\
\vdots\\
x(N)
\end{bmatrix},
\quad
U =
\begin{bmatrix}
u(0)\\
u(1)\\
\vdots\\
u(N-1)
\end{bmatrix}
\end{equation}

The prediction model is

\begin{equation}
X = \bar{A}x(0) + \bar{B}U
\end{equation}

where $\bar{A}$ and $\bar{B}$ are the standard stacked prediction matrices constructed from $A_d$ and $B_d$.

\subsection{Cost Function Formulation}

The finite-horizon objective is defined as

\begin{equation}
J =
\frac{1}{2}\sum_{k=1}^{N} x(k)^TQx(k)
+
\frac{1}{2}\sum_{k=0}^{N-1} u(k)^TRu(k)
+
\frac{1}{2}x(0)^TQx(0)
\end{equation}

where $Q$ penalizes state deviation and $R$ penalizes actuator effort.

Using

\begin{equation}
X = \bar{A}x(0) + \bar{B}U
\end{equation}

the cost becomes

\begin{equation}
J(x(0),U)
=
\text{constant}(x(0))
+
\frac{1}{2}U^THU
+
x(0)^TF^TU
\end{equation}

where

\begin{equation}
H = \bar{B}^{T}\bar{Q}\bar{B} + \bar{R}
\end{equation}

\begin{equation}
F^{T} = \bar{A}^{T}\bar{Q}\bar{B}
\end{equation}

with

\begin{equation}
\bar{Q} = \operatorname{diag}(Q,Q,\ldots,Q),
\quad
\bar{R} = \operatorname{diag}(R,R,\ldots,R)
\end{equation}

\subsection{Terminal Constraint Formulation}

Terminal convergence is enforced through

\begin{equation}
x(N) = 0
\end{equation}

Using the prediction model,

\begin{equation}
x(N) = A_Nx(0) + B_NU
\end{equation}

yielding the terminal constraint

\begin{equation}
B_NU = -A_Nx(0)
\end{equation}

\subsection{Optimal Control Law}

The constrained optimization problem is formulated as

\begin{equation}
\min_U
\frac{1}{2}U^THU + x(0)^TF^TU
\end{equation}

subject to

\begin{equation}
u_{\min} \leq U \leq u_{\max}
\end{equation}

and

\begin{equation}
B_NU = -A_Nx(0)
\end{equation}

The resulting quadratic program is solved using MATLAB \texttt{quadprog} to obtain the optimal control sequence

\begin{equation}
U^{*} =
\begin{bmatrix}
u^{*}(0) & u^{*}(1) & \cdots & u^{*}(N-1)
\end{bmatrix}^{T}
\end{equation}

\subsection{Baseline Trajectory Planning Methods}

The proposed framework is compared against cubic, quintic, and trapezoidal trajectory planners defined respectively as

\begin{equation}
q(t) = a_0 + a_1t + a_2t^2 + a_3t^3
\end{equation}

\begin{equation}
q(t) = a_0 + a_1t + a_2t^2 + a_3t^3 + a_4t^4 + a_5t^5
\end{equation}

and

\begin{equation}
q(t) =
\begin{cases}
q_s + \frac{1}{2}\ddot{q}_bt^2, & 0 \leq t < t_b\\
q_s + \dot{q}_b\left(t-\frac{t_b}{2}\right), & t_b \leq t < t_g - t_b\\
q_g - \frac{1}{2}\ddot{q}_b(t_g - t)^2, & t_g - t_b \leq t \leq t_g
\end{cases}
\end{equation}

Unlike the proposed framework, these methods do not explicitly incorporate manipulator dynamics or actuator control effort during trajectory generation.

\section{Experimental Setup}

Experiments were conducted using a reduced 3-DoF UR5 robotic manipulator operating under nonlinear rigid-body dynamics. A point-to-point joint-space motion task was considered over a fixed trajectory duration. All simulations used $T_f=2$ s, $\Delta t=0.02$ s, $N=100$, identical PID gains, corrective torque saturation $|\Delta\tau_i|\leq 60$ Nm, and the same nonlinear RK4 execution model for all planners.

The proposed control-aware optimal trajectory planner was compared against cubic polynomial, quintic polynomial, and trapezoidal velocity-profile trajectory planners. All methods were evaluated under identical initial and terminal conditions, sampling time, actuator constraints, and nonlinear execution settings to ensure fair comparison. To isolate the influence of trajectory generation on nonlinear execution behavior, all planned trajectories were executed using the same nonlinear manipulator model and identical feedforward-assisted PID tracking controller. The applied control input was defined as

\begin{equation}
\Delta \tau_{cmd}
=
\Delta \tau_{ff}
+
\Delta \tau_{fb}
\end{equation}

where the feedback control term was given by

\begin{equation}
\Delta \tau_{fb}
=
K_p e
+
K_i \int e \, dt
+
K_d \dot{e}
\end{equation}

with tracking error

\begin{equation}
e = q_{ref} - q
\end{equation}

Trajectory planners were evaluated using tracking and corrective-control metrics computed during nonlinear closed-loop execution. Tracking accuracy was evaluated using RMS tracking error, while corrective actuator demand was evaluated using RMS corrective torque and cumulative corrective-control activity. Overall execution performance was further evaluated using the cumulative executed cost

\begin{equation}
J_{exec}
=
\sum_{k=0}^{N-1}
\left(
x_k^T Q_{exec} x_k
+
u_k^T R_{exec} u_k
\right)
\end{equation}

where \(Q_{exec}\) penalizes tracking deviation and \(R_{exec}\) penalizes corrective control effort during nonlinear execution.

Corrective actuator demand was evaluated using RMS corrective torque and the cumulative corrective-control activity

\begin{equation}
J_{\Delta \tau}
=
\int_0^T
\sum_i
|\Delta \tau_i(t)| \, dt
\end{equation}

which quantifies the total corrective actuator effort required during nonlinear execution.

\section{Results and Discussion}

All trajectory planners were evaluated under identical nonlinear execution conditions using the same nonlinear UR5 manipulator model, feedforward-assisted PID tracking controller, controller gains, actuator constraints, and simulation settings, ensuring that observed performance differences arise from the trajectory generation method itself.

\subsection{Nonlinear Trajectory Execution}

\begin{figure}[!t]
\centering
\includegraphics[width=\linewidth]{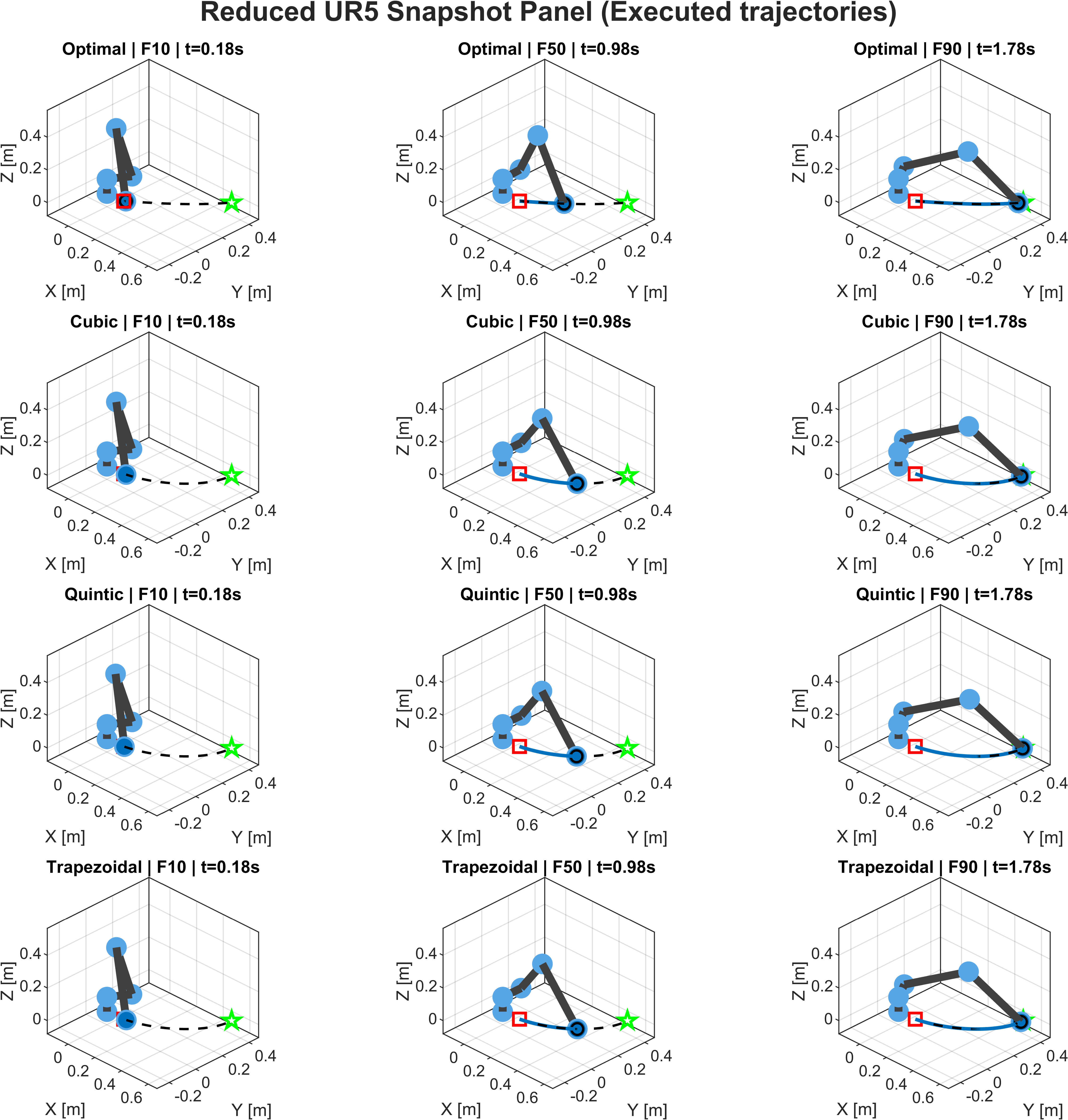}
\caption{Representative nonlinear execution snapshots.}
\label{fig:nonlinear_execution}
\end{figure}

Fig.~\ref{fig:nonlinear_execution} presents representative nonlinear execution snapshots for the evaluated trajectory planners under identical execution conditions. Although all methods satisfy the same point-to-point motion objective, noticeable differences can be observed in the resulting execution behavior throughout trajectory evolution.

\subsection{Planned Trajectory Characteristics}

\begin{figure}[!t]
\centering
\includegraphics[width=\linewidth]{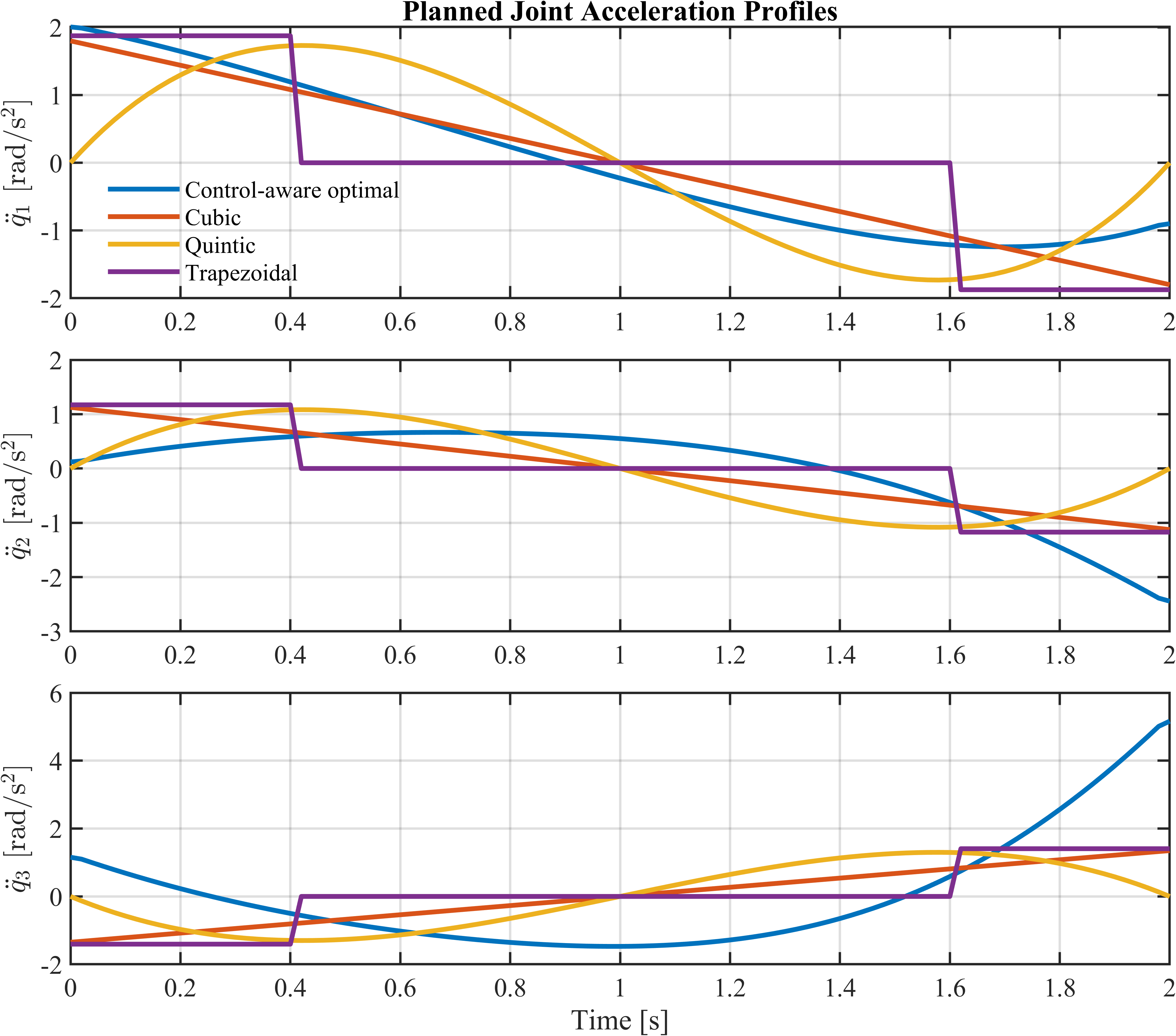}
\caption{Planned joint acceleration profiles.}
\label{fig:acceleration_profiles}
\end{figure}

Fig.~\ref{fig:acceleration_profiles} compares the planned joint acceleration profiles generated by the evaluated trajectory planners. The trapezoidal planner produces abrupt acceleration transitions due to its piecewise velocity-profile structure, while the quintic planner exhibits comparatively large acceleration variations despite its higher-order smoothness formulation. In contrast, the proposed control-aware optimal planner produces smoother and more gradually distributed acceleration behavior across all joints, indicating improved compatibility with nonlinear manipulator dynamics during execution.

\subsection{Tracking Performance Under Nonlinear Execution}

\begin{figure}[!t]
\centering
\includegraphics[width=\linewidth]{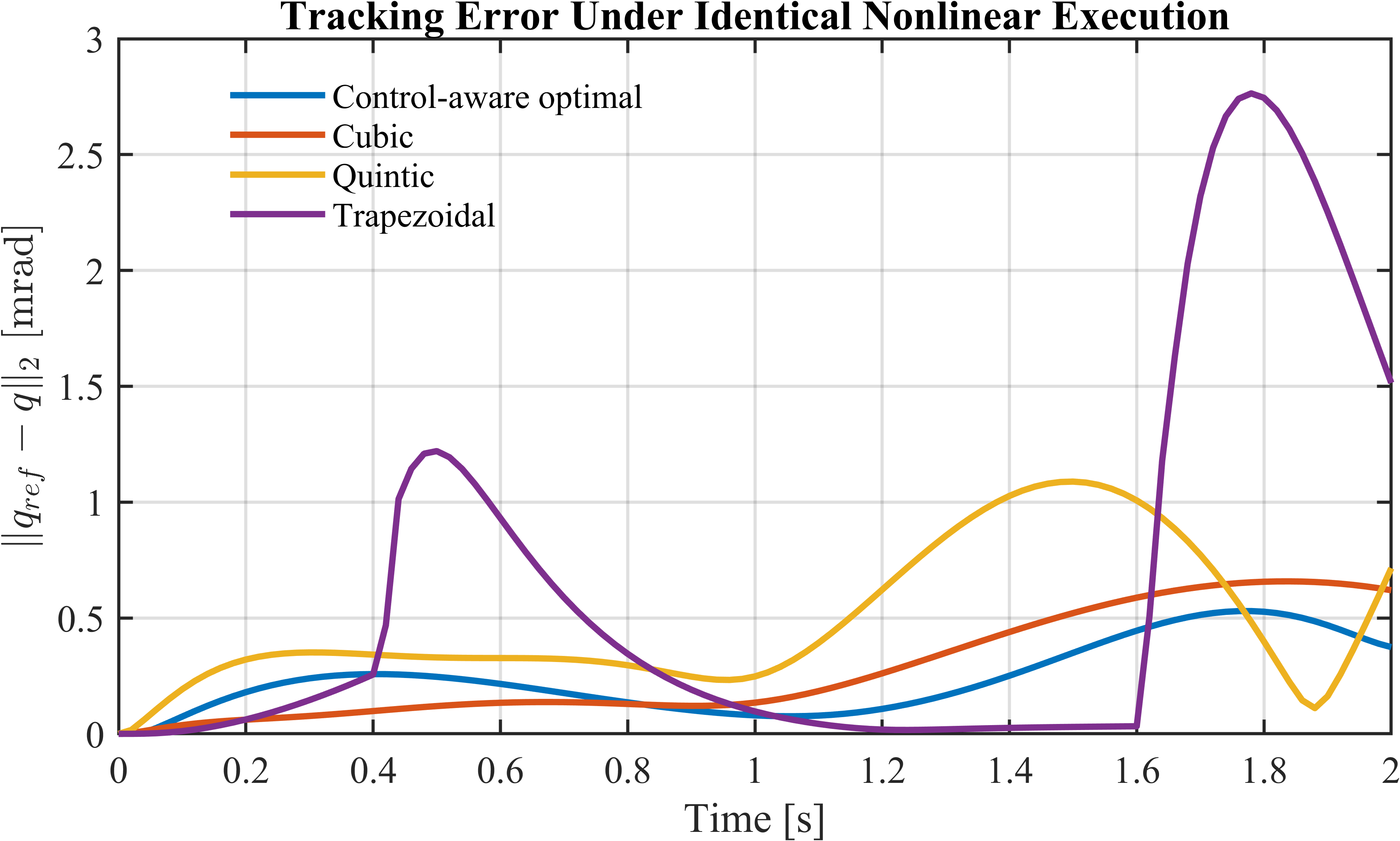}
\caption{Tracking error comparison.}
\label{fig:tracking_error}
\end{figure}

Fig.~\ref{fig:tracking_error} compares the closed-loop tracking error under nonlinear execution. The proposed control-aware optimal planner achieved the lowest tracking error throughout most of the trajectory execution, while the trapezoidal planner exhibited the largest tracking deviations near acceleration transition regions.

Quantitatively, the proposed method achieved an RMS tracking error of \(2.825\times10^{-4}\,\mathrm{rad}\), compared to \(3.622\times10^{-4}\,\mathrm{rad}\), \(5.653\times10^{-4}\,\mathrm{rad}\), and \(1.059\times10^{-3}\,\mathrm{rad}\) for the cubic, quintic, and trapezoidal planners, respectively. Relative to the proposed method, RMS tracking error increased by approximately 22\%, 50\%, and 73\% for the cubic, quintic, and trapezoidal planners.

\subsection{Corrective Control Effort Analysis}

\begin{figure}[!t]
\centering
\includegraphics[width=\linewidth]{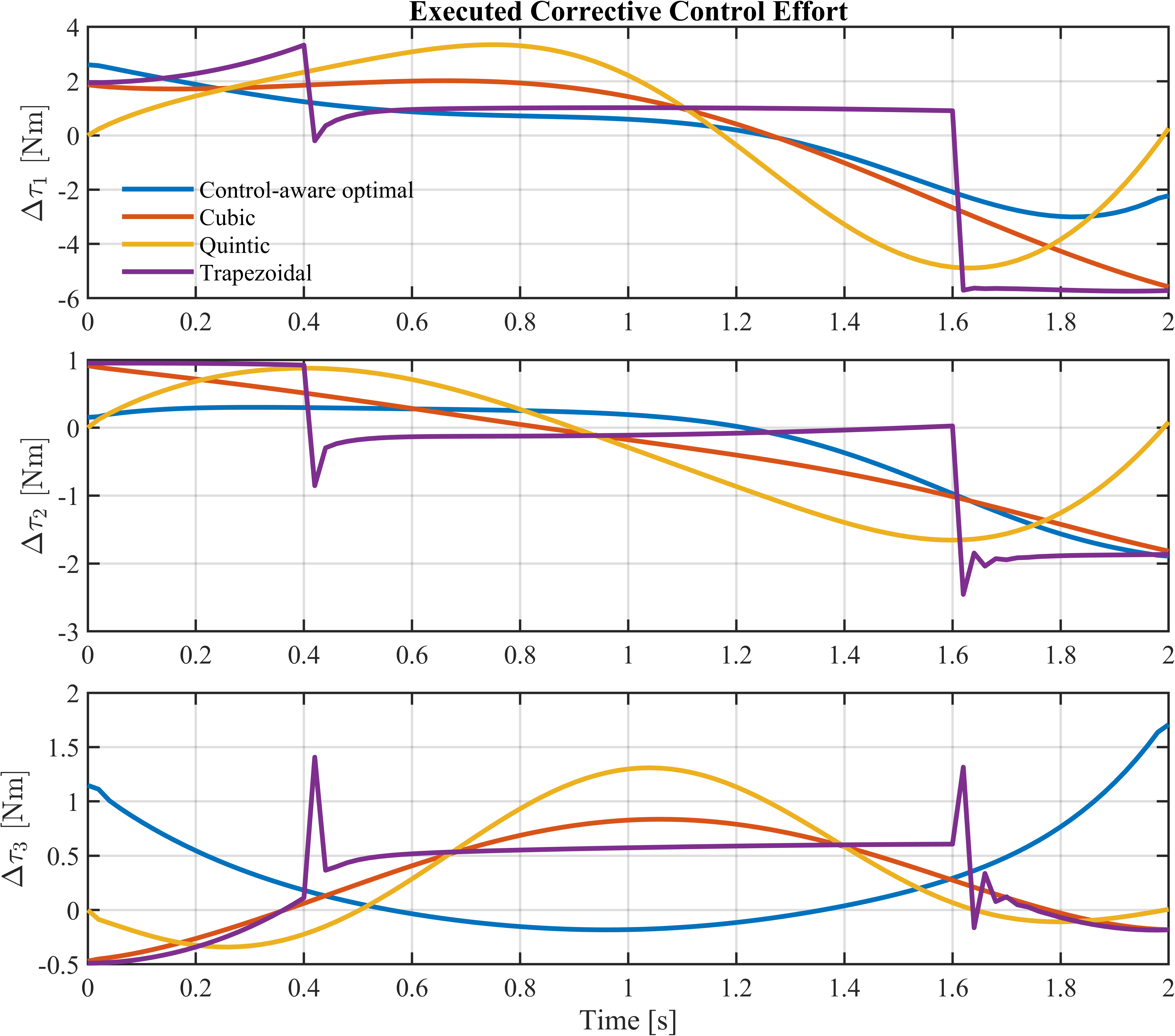}
\caption{Corrective control effort.}
\label{fig:corrective_effort}
\end{figure}

Fig.~\ref{fig:corrective_effort} compares the corrective actuator commands required during nonlinear execution. The proposed control-aware optimal planner consistently required lower corrective control effort compared to the classical trajectory planners.

\begin{figure}[!t]
\centering
\includegraphics[width=\linewidth]{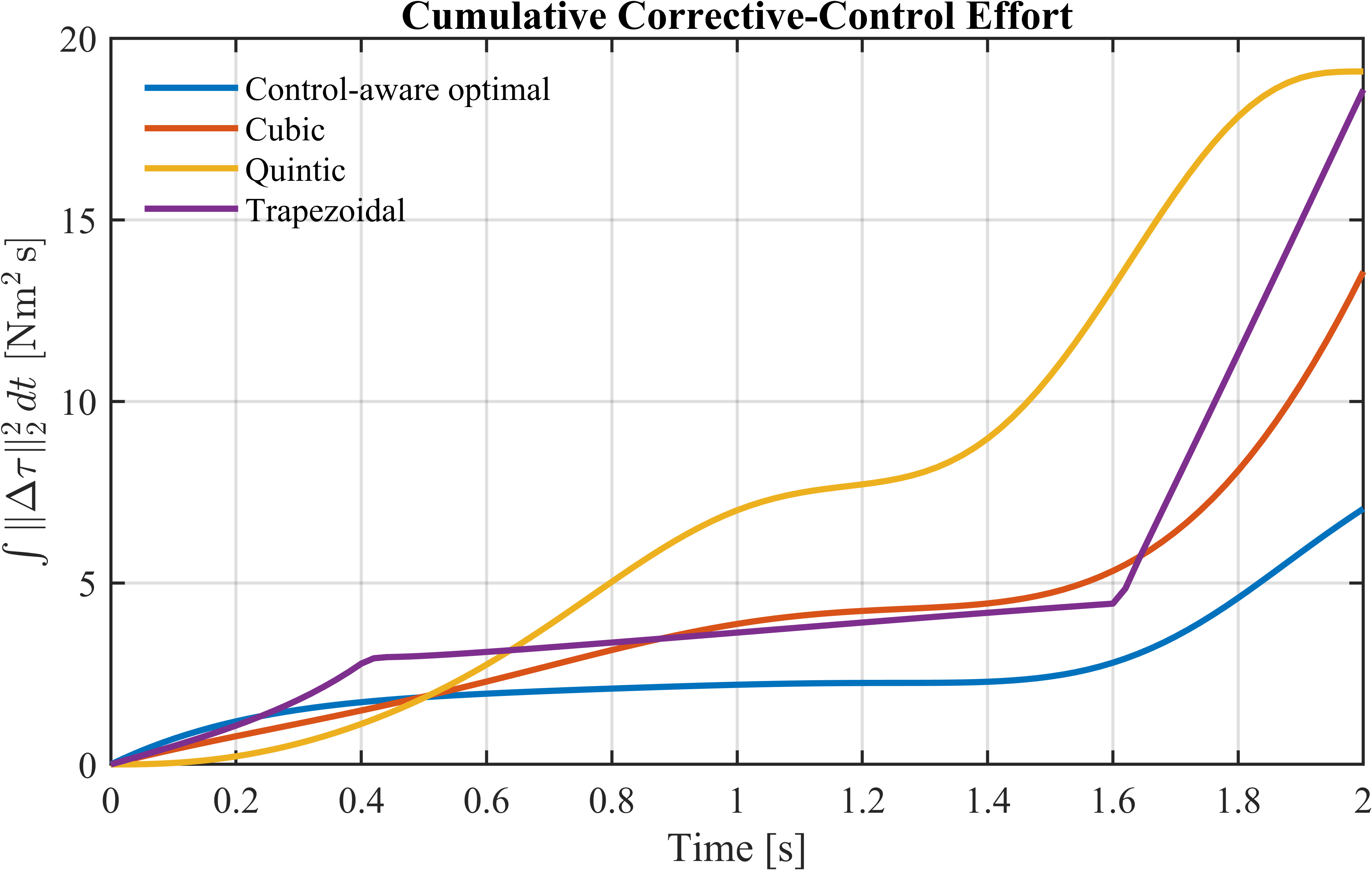}
\caption{Cumulative corrective-control activity.}
\label{fig:cumulative_activity}
\end{figure}

Cumulative corrective-control activity was evaluated using

\begin{equation}
J_{\Delta \tau} =
\int_{0}^{T}
\sum_i
\left|
\Delta \tau_i(t)
\right|
\, dt
\end{equation}

As shown in Fig.~\ref{fig:cumulative_activity}, the proposed planner produced the lowest cumulative corrective-control activity among all compared methods.

The proposed method achieved an RMS corrective torque of \(1.893\,\mathrm{N\,m}\), compared to \(2.629\,\mathrm{N\,m}\), \(3.074\,\mathrm{N\,m}\), and \(3.066\,\mathrm{N\,m}\) for the cubic, quintic, and trapezoidal planners, respectively. Similarly, the integrated corrective-control activity was \(4.535\,\mathrm{N\,m\,s}\), compared to \(6.329\,\mathrm{N\,m\,s}\), \(7.723\,\mathrm{N\,m\,s}\), and \(6.504\,\mathrm{N\,m\,s}\), respectively. Relative to the classical planners, the proposed method reduced corrective actuator demand by approximately 28--41\%.

\subsection{Closed-Loop Execution Cost Comparison}

\begin{figure}[!t]
\centering
\includegraphics[width=\linewidth]{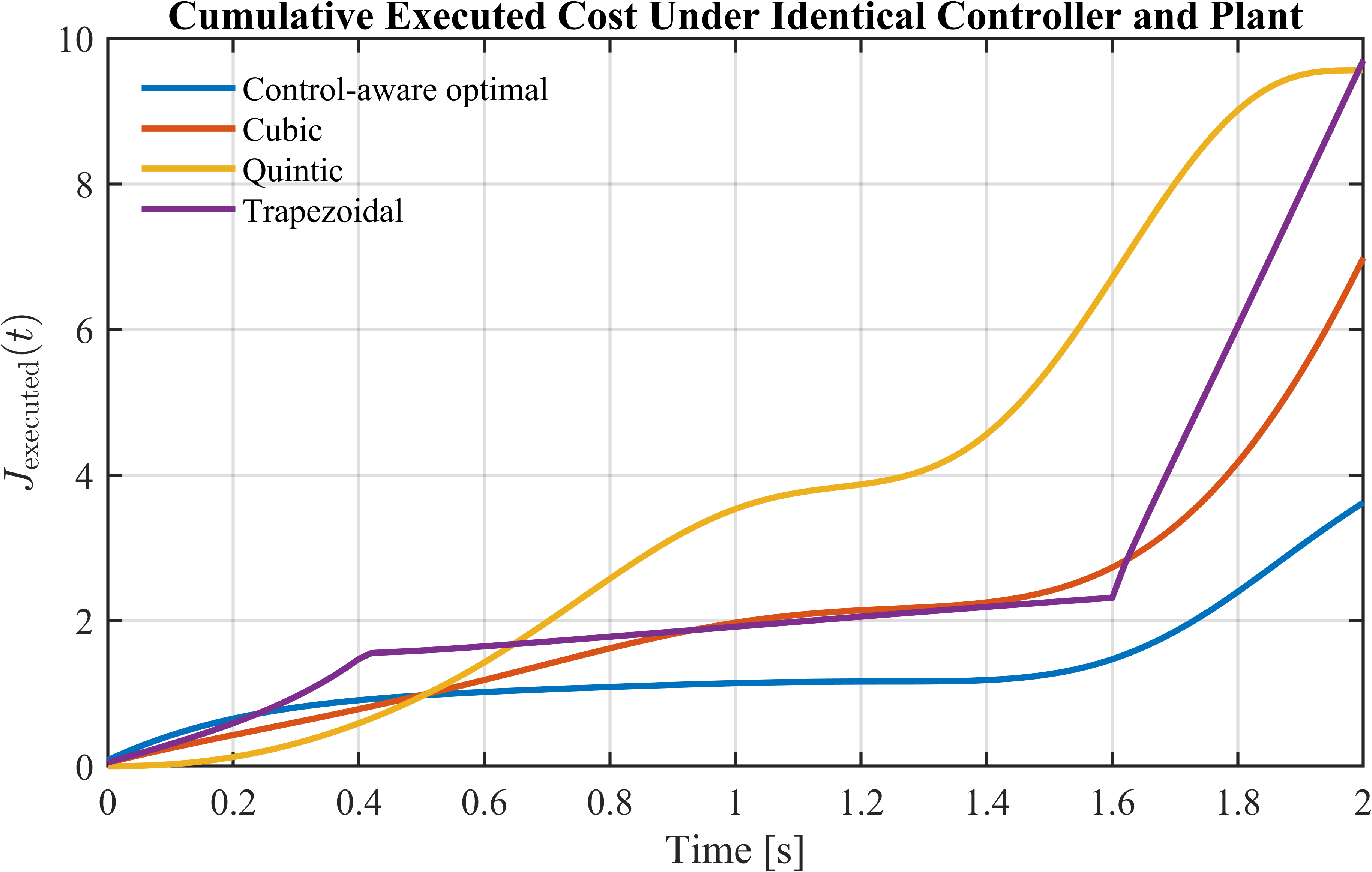}
\caption{Cumulative executed cost.}
\label{fig:executed_cost}
\end{figure}

Fig.~\ref{fig:executed_cost} compares the cumulative executed cost accumulated during nonlinear trajectory execution. The proposed control-aware planner consistently maintained the lowest executed cost throughout the motion duration.

The proposed method achieved a final executed cost of \(3.622\), compared to \(6.983\), \(9.562\), and \(9.697\) for the cubic, quintic, and trapezoidal planners, respectively. Relative to the classical planners, the proposed method reduced executed cost by approximately 48\% relative to the cubic planner and more than 62\% relative to both the quintic and trapezoidal planners.

\subsection{Quantitative Performance Summary}

\begin{table}[!t]
\caption{Quantitative performance comparison}
\label{tab:performance_metrics}
\centering
\scriptsize
\renewcommand{\arraystretch}{1.05}
\resizebox{\columnwidth}{!}{%
\begin{tabular}{lcccccc}
\hline
Planner & RMS $\Delta\tau$ & Int. $\Delta\tau$ & RMS Err. & Cost & $\Delta\tau$ Red. & Cost Red. \\
& N\,m & N\,m\,s & rad & -- & \% & \% \\
\hline
Optimal & 1.893 & 4.535 & $2.825{\times}10^{-4}$ & 3.622 & -- & -- \\
Cubic & 2.629 & 6.329 & $3.622{\times}10^{-4}$ & 6.983 & 28.35 & 48.13 \\
Quintic & 3.074 & 7.723 & $5.653{\times}10^{-4}$ & 9.562 & 41.29 & 62.12 \\
Trap. & 3.066 & 6.504 & $1.059{\times}10^{-3}$ & 9.697 & 30.27 & 62.64 \\
\hline
\end{tabular}%
}
\end{table}

\begin{figure}[!t]
\centering
\includegraphics[width=\linewidth]{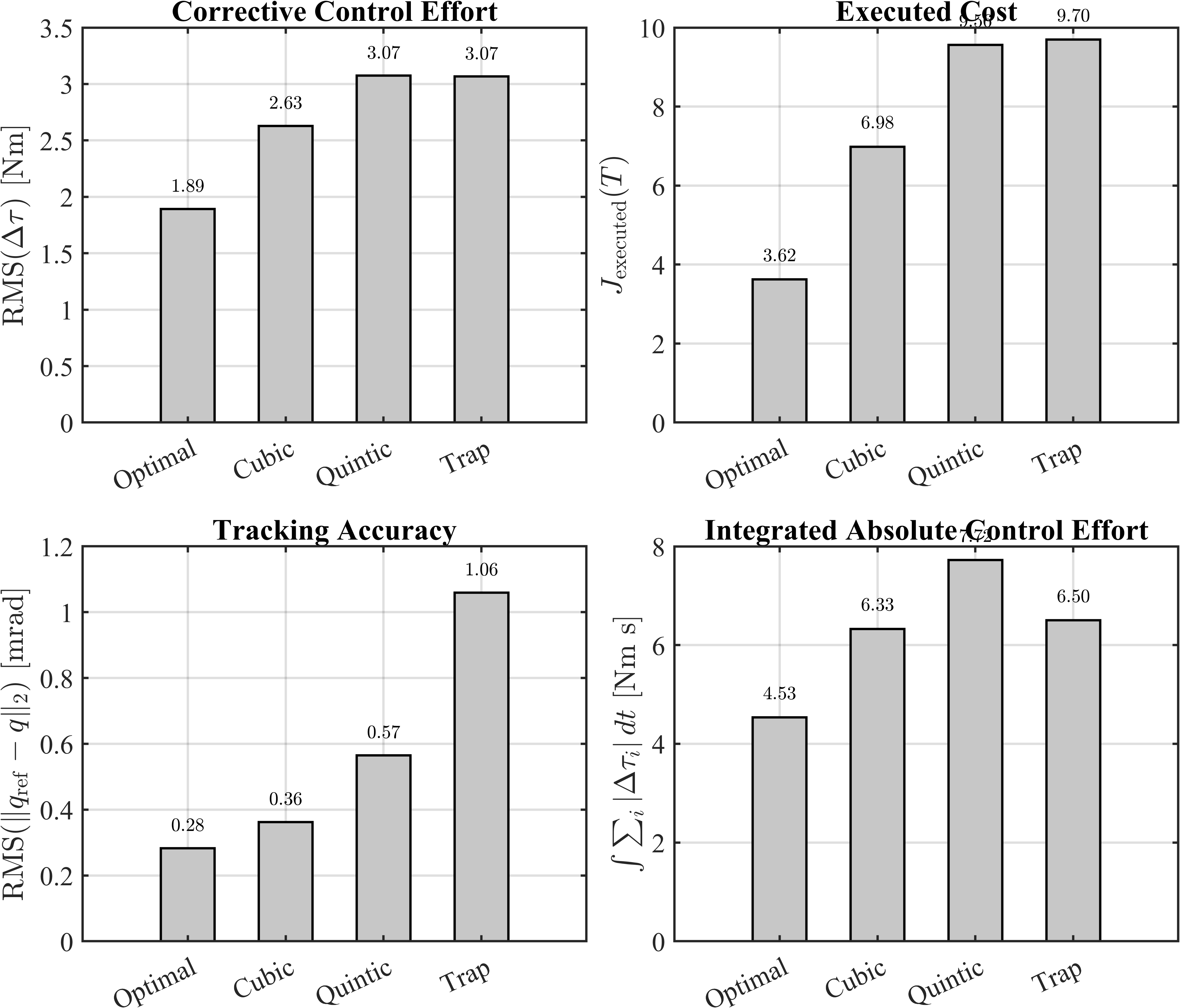}
\caption{Quantitative performance comparison.}
\label{fig:quantitative_comparison}
\end{figure}

Table~\ref{tab:performance_metrics} and Fig.~\ref{fig:quantitative_comparison} summarize the quantitative performance metrics across all evaluated trajectory planners. Overall, the results demonstrate that although classical trajectory planners provide smooth kinematic trajectories, they can produce substantially higher corrective actuator demand and execution cost under nonlinear manipulator dynamics. Across all evaluated metrics, the proposed control-aware optimal planner consistently achieved superior nonlinear execution performance.

\section{Conclusion}
This paper investigated the nonlinear execution behavior of classical kinematic trajectory planners and compared them against a proposed control-aware optimal trajectory planning framework for robotic manipulators. Unlike conventional trajectory planners that primarily emphasize kinematic smoothness, the proposed method explicitly incorporated actuator control effort and manipulator dynamics within a finite-horizon optimal control formulation. A unified nonlinear evaluation framework was further developed in which all trajectory planners were executed under identical nonlinear dynamics, controller configurations, and actuator constraints to enable fair comparison of execution behavior.
Simulation results using a nonlinear 3-DoF UR5 manipulator demonstrated that the proposed method consistently reduced tracking error, corrective actuator demand, cumulative corrective-control activity, and closed-loop execution cost compared to cubic, quintic, and trapezoidal trajectory planners. The results further showed that higher-order kinematic smoothness alone does not necessarily correspond to dynamically efficient nonlinear manipulator execution.


\begin{thebibliography}{99}


\bibitem{b1} B. Siciliano, L. Sciavicco, L. Villani, and G. Oriolo, \textit{Robotics: Modelling, Planning and Control}. London, U.K.: Springer, 2009.
\bibitem{b2} A. A. Ata, ``Optimal trajectory planning of manipulators: A review,'' \textit{Journal of Engineering Science and Technology}, vol. 2, no. 1, pp. 32--54, Apr. 2007.
\bibitem{b3} Z. Wu, J. Chen, D. Zhang, J. Wang, L. Zhang, and F. Xu, ``A novel multi-point trajectory generator for robotic manipulators based on piecewise motion profile and series-parallel analytical strategy,'' \textit{Mechanism and Machine Theory}, vol. 181, Art. no. 105201, Mar. 2023.
\bibitem{b4} R. Hedjar and P. Boucher, ``Nonlinear receding-horizon control of rigid link robot manipulators,'' \textit{International Journal of Advanced Robotic Systems}, vol. 2, no. 1, pp. 21--30, Mar. 2005.
\bibitem{b5} S. Chen and J. T. Wen, ``Neural-learning trajectory tracking control of flexible-joint robot manipulators with unknown dynamics,'' arXiv preprint arXiv:1908.03269, Aug. 2019.
\bibitem{b6} M. Kelly, ``An introduction to trajectory optimization: How to do your own direct collocation,'' \textit{SIAM Review}, vol. 59, no. 4, pp. 849--904, 2017.
\bibitem{b7} J. Wilson, M. Charest, and R. Dubay, ``Non-linear model predictive control schemes with application on a 2 link vertical robot manipulator,'' \textit{Robotics and Computer-Integrated Manufacturing}, vol. 41, pp. 23--30, Oct. 2016.
\bibitem{b8} F. Rubio, F. Valero, J. Sunyer, and J. Cuadrado, ``Optimal time trajectories for industrial robots with torque, power, jerk and energy consumed constraints,'' \textit{Industrial Robot: An International Journal}, vol. 39, no. 1, pp. 92--100, 2012.
\bibitem{b9} A. Gasparetto, P. Boscariol, A. Lanzutti, and R. Vidoni, ``Path planning and trajectory planning algorithms: A general overview,'' in \textit{Motion and Operation Planning of Robotic Systems: Background and Practical Approaches}, G. Carbone and F. Gomez-Bravo, Eds. Cham, Switzerland: Springer, 2015, pp. 3--27.
\bibitem{b10} S. R. Munasinghe, \textit{Optimization and Optimal Control in a Nutshell}. Singapore: Springer Nature Singapore, 2024.
\bibitem{b11} W. Li and E. Todorov, ``Iterative linear quadratic regulator design for nonlinear biological movement systems,'' in \textit{Proc. 1st Int. Conf. Informatics in Control, Automation and Robotics (ICINCO)}, Setúbal, Portugal, Aug. 2004, pp. 222--229.
\bibitem{b12} A. Gasparetto and V. Zanotto, ``A technique for time-jerk optimal planning of robot trajectories,'' \textit{Robotics and Computer-Integrated Manufacturing}, vol. 24, no. 3, pp. 415--426, Jun. 2008.
\bibitem{b13} P. Triantafyllou, H. Mnyusiwalla, P. Sotiropoulos, M. A. Roa, D. Russell, and G. E. Deacon, ``A benchmarking framework for systematic evaluation of robotic pick-and-place systems in an industrial grocery setting,'' in \textit{Proc. IEEE Int. Conf. Robotics and Automation (ICRA)}, Montreal, QC, Canada, 2019, pp. 6692--6698.

\end{thebibliography}
\end{document}